\documentclass[reprint,amsmath,amssymb,aps]{revtex4-2}

\usepackage{kpfonts}
\AtBeginDocument{\selectfont} 
\usepackage[dvipsnames]{xcolor}
\usepackage[english]{babel}
\usepackage{graphicx}
\usepackage{dcolumn}
\usepackage{bm}
\usepackage[caption=false]{subfig}
\usepackage[colorlinks=true, linkcolor=blue, citecolor=blue, urlcolor=blue]{hyperref}
\usepackage[utf8]{inputenc}
\usepackage{graphicx}
\usepackage{float}             
\usepackage[export]{adjustbox} 
\makeatletter
\providecommand{\l@en}{\l@english} 
\makeatother
\begin{document}


\title{Physical Complexity of a Cognitive Artifact}

\author{Gülce Kardeş}
\affiliation{Department of Computer Science, University of Colorado Boulder, Boulder, CO, USA}
\affiliation{Santa Fe Institute, Santa Fe, NM, USA}

\author{David Krakauer}
\affiliation{Santa Fe Institute, Santa Fe, NM, USA}

\author{Joshua A. Grochow}
\affiliation{Department of Computer Science, University of Colorado Boulder, Boulder, CO, USA}
\affiliation{Department of Mathematics, University of Colorado Boulder, Boulder, CO, USA}

\date{September 9, 2025}

\begin{abstract}
Cognitive science and theoretical computer science both seek to classify and explain the difficulty of tasks. Mechanisms of intelligence can be understood as leading to reductions in task difficulty. We map concepts from  the computational complexity of a physical puzzle, the Soma Cube, onto cognitive problem-solving strategies through a ``Principle of Materiality''.  By analyzing the puzzle's branching factor, measured through search tree outdegree, we quantitatively assess task difficulty and systematically examine how different strategies modify complexity. We incrementally enhance a trial-and-error tree search with preprocessing (cognitive chunking), value ordering (cognitive free-sorting), variable ordering (cognitive scaffolding), and pruning (cognitive inference). We discuss how the competent use of artifacts reduces time complexity through the exploitation of physical constraints, and propose a model of intelligence as a library of algorithms exploiting the capabilities of both mind and matter.
\end{abstract}
\maketitle

\section{\label{sec:level1}Introduction}
There are currently two well established domains for studying general problem solving. The first describes strategies used by humans on both experimental and real-world tasks. Human problem solving is captured through a number of frameworks including skill acquisition \cite{Haith2018-om} and automaticity \cite{Krakauer2020-wg}, the application of expert knowledge \cite{Ericsson1993-cs, Charness2008-nq}, the use of heuristics \cite{Gigerenzer2021-jy, Chronicle2004-gb}, reinforcement learning and conditioning \cite{Sutton1998-wd, Botvinick2019-nc}, Bayesian inference \cite{Ullman2020-wt, Griffiths2023-my}, analogy making \cite{Holyoak2001-di, Hofstadter2001-eq}, collective intelligence and cognition \cite{Couzin2009-kr, dreyerEtAlFeinerman}, simulation intelligence \cite{Lavin2021-jr}, the use of external representations \cite{Norman1975-pd, Zhang1997-qi} and the synergy of mind and matter through exbodiment \cite{krakauer2024exbodimentmindmatter}. 

The second domain, computational problem solving, investigates algorithms that enable computers to tackle problems effectively. Within this domain, two branches especially pertinent to the present question are: (i) computational complexity theory, which analyzes the resources (time, memory, etc.) required to solve problems as functions of input size, typically in the asymptotic limit; and (ii) the study of search algorithms, which seeks efficient solutions to specific tasks (e.g., games and puzzles) by exploiting the combinatorial structure of state spaces, often via heuristics \cite{10.5555/1882723.1882745, knuth2011art, Anstegui2010, kociemba2025}.

Although both humans and computers have been shown to be capable of solving a large set of common tasks \cite{Das2015-zt, Mandal2015-fr}, and there is an extensive literature comparing natural and artificial intelligence \cite{Korteling2021-vz, Griffiths2023-my, Mitchell2019-fg}, comparing human and computer problem-solving remains challenging.  One problem is that  algorithms can be viewed as ``white boxes'' designed by humans to solve problems efficiently, whereas human problem solving is a ``black box'' evolved (or learned) to solve problems, that can only be probed indirectly. Even with carefully constructed benchmarks, it is not at all obvious whether humans and computers are really solving the same problems \cite{McCarthy1981-aa}. 

Here, we present a new approach to the challenge of making analogies between these two approaches 
by mapping formal properties of search algorithms onto the existing literature on experimentally observed mechanisms of cognition. We seek to establish a common language for well-established phenomena from both problem-solving domains. Using narrowly defined problems \cite{Mechner2010-bx} or benchmarks \cite{Chollet2019-te}, which retain sufficient richness to permit diverse solutions but whose configuration space is modest enough to be amenable to analysis, we set out on a path towards a convergence of these perspectives. Our crucible for accomplishing this goal will be the analysis of algorithmic puzzle-solving applied to a physical artifact, the Soma Cube. 
\section{Physical artifact and logical puzzle: the Soma Cube and the Principle of Materiality}
\label{sec:POM}
\begin{figure}[h!]
        \centering
        \includegraphics[width=\columnwidth]{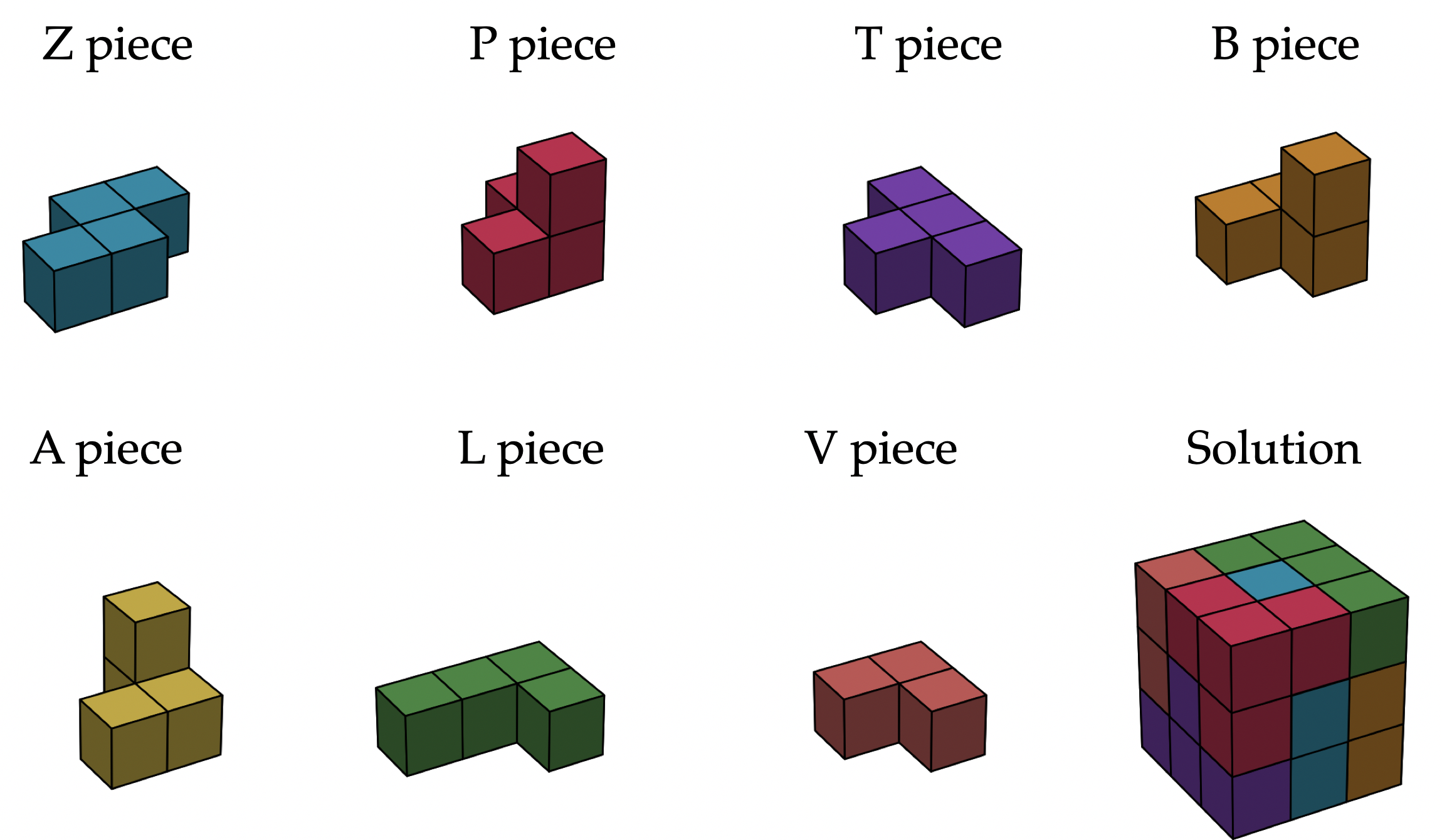}  
        \caption{ \label{fig:cube} The seven pieces of the Soma Cube. In the original puzzle these are not colored and color plays no part in the final solution.  We use a consistent coloring scheme throughout for expositional purposes.  Also shown is one of the 240 solutions of the Soma cube: a $3 \times 3 \times 3$ cube with no gaps or “voids” in the structure.
        }
\end{figure}
The Soma Cube is a 3D puzzle made of seven distinct polycubic pieces.  The goal is to assemble the pieces into a $3 \times 3 \times 3$ cube with no gaps (Fig.~\ref{fig:cube}). There are 240 essentially distinct solutions to the Soma Cube, treating solutions that are rotations or reflections of one another as equivalent \cite{Berlekamp_2004}. 

\begin{figure}[h!]
        \centering
\includegraphics[width=\columnwidth]{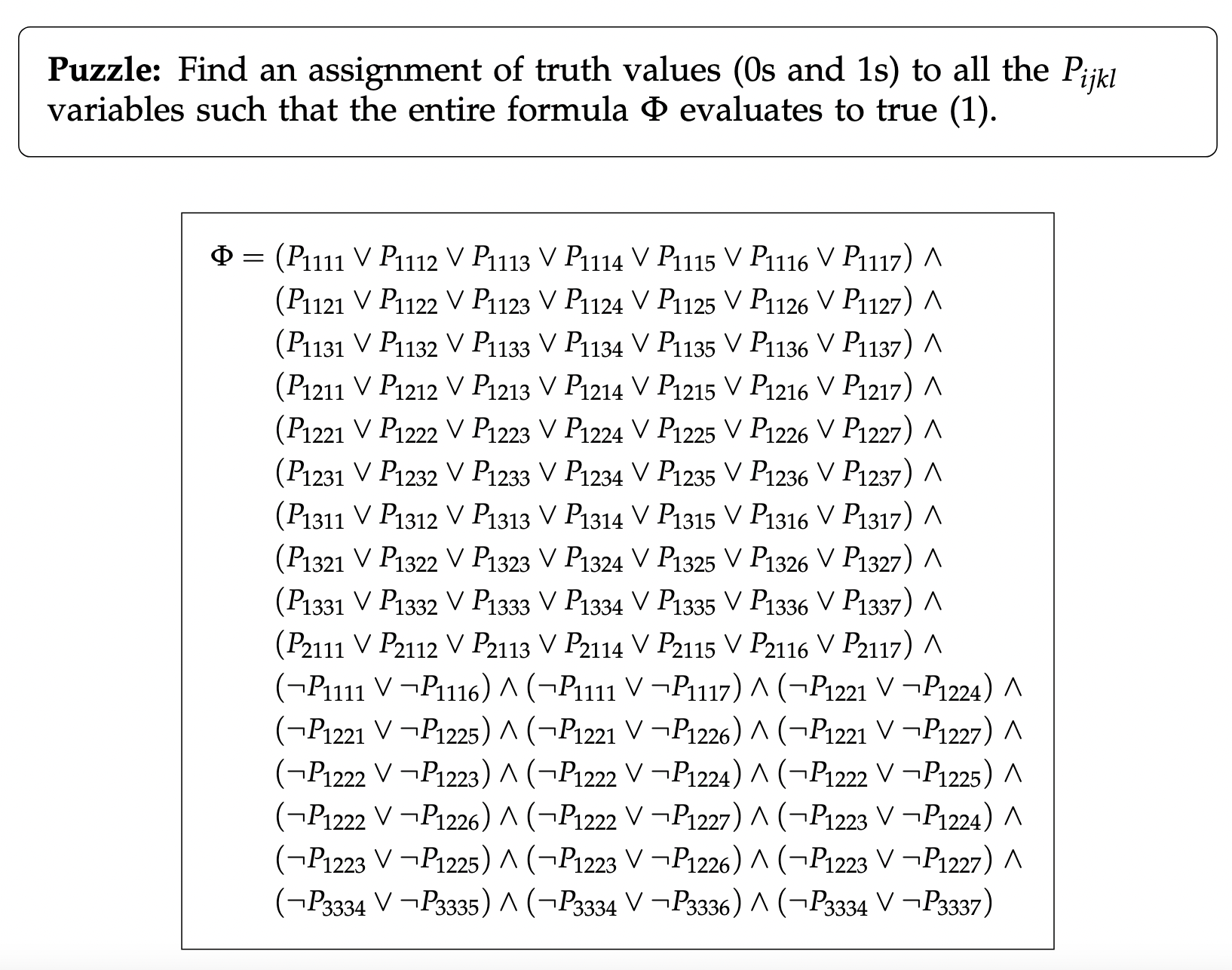}  
        \caption{ \label{fig:SAT} A fragment of the k-SAT formulation of the Soma Cube puzzle. The variables $P_{ijkl}$ represent whether piece $l$ occupies position $(i,j,k)$ in the $3\times3\times3$ cube. Although only a small portion is shown here, the complete formulation consists of over 150,000 clauses that enforce both piece placement and cell coverage constraints, ensuring that every piece is used exactly once and every cell in the cube is occupied by one piece.} 
\end{figure}

The puzzle also serves as a physical realization of an abstract mathematical problem, which can be expressed in a variety of different formalisms. For example, there are various Boolean formulae whose satisfying assignments correspond to solutions of the Soma Cube, one of which is shown in part in Fig.~\ref{fig:SAT}---an instance of the Boolean Satisfiability (SAT) problem. The complete formula $\Phi$ consists of constraints that encode the valid placements and orientations of the seven pieces within the cube (the details of this encoding can be found in App. \ref{sec:SI1}). 

The physical Soma Cube and its abstract SAT formulation are in some sense representations of the same puzzle. However, human solvers find the physical puzzle to be tractable, whereas the corresponding SAT puzzle with more than $10^5$ clauses is intractable for humans to solve through direct reasoning. This stark contrast in solvability underscores a broader principle: sensitivity to problem representation. The way a problem is framed shapes its state space, and the combinatorial properties of that space ultimately determine its complexity, whether for human cognition, AI, or other algorithms \cite{Amarel1968OnRO}.

We represent the state space of the Soma Cube as a configuration graph $G_P = (V_P, E_P)$, where the vertices $V_P$ encode distinct piece configurations, and the edges $E_P = \{(u,v) : u,v \in V_P\}$ encode legal single-step moves between states (adding or removing a piece). Most natural solvers of the Soma Cube (beyond naive enumeration) operate on a search space graph $G_S = (V_S, E_S)$ that is a subgraph of the full configuration graph $G_P$ \footnote{A solver could in principle operate by putting together several multi-piece configurations independently before combining them, but we leave analysis of such solvers to future work.}. The efficiency of a solver depends largely on the structure and size of the search space it explores, as the combinatorial properties of \( G_S \) directly impact computational complexity. Analyzing the combinatorial properties of the state space graph uncovers determinants of the problem’s difficulty and informs the algorithms best suited to solve it (see App. \ref{sec:SI2}).

\begin{figure}[h!]
  \centering

  \subfloat[Average branching factor decay across depths for the Soma Cube puzzle, showing exponential reduction in possible moves as pieces are placed.]{
    \includegraphics[width=\columnwidth]{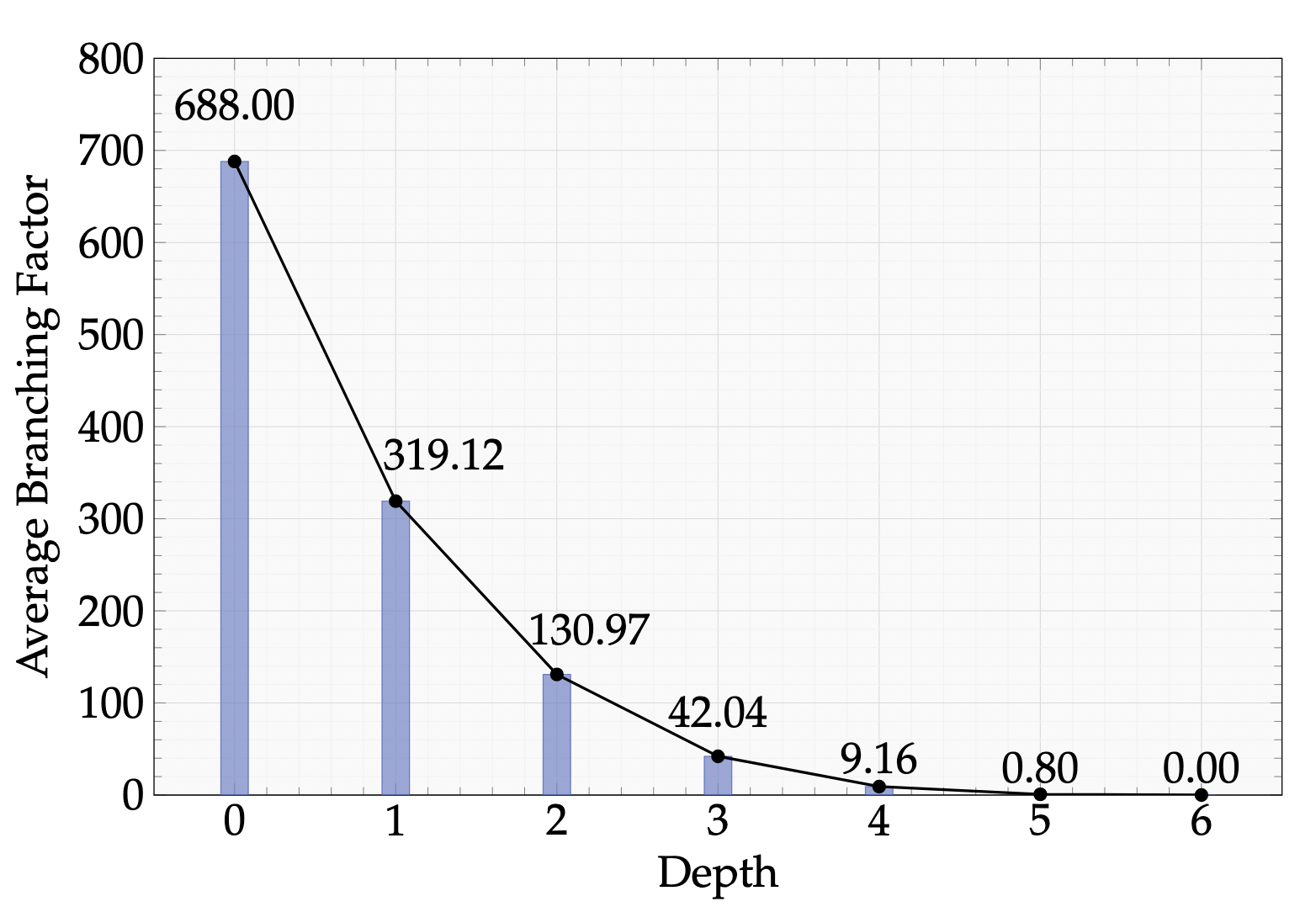}
  }

  \vspace{0.8em}

  \subfloat[Distribution of branching factors at different depths (1--6) in the Soma Cube puzzle state space.]{
    \includegraphics[width=\columnwidth]{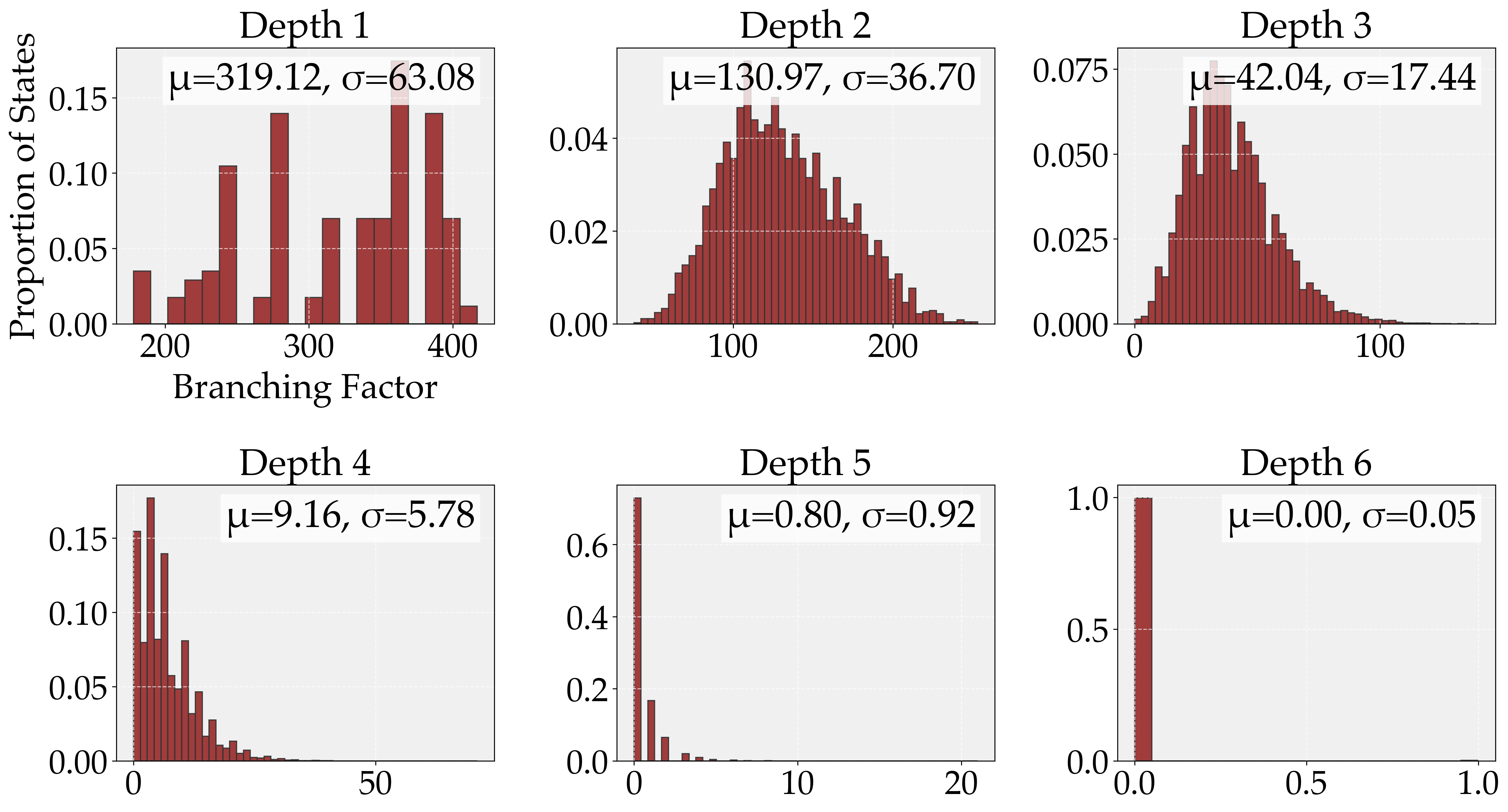}
  }

  \caption{Branching factors in the Soma Cube puzzle: (a) The average shows an exponential decay as the depth increases. This is characteristic of puzzles where each new move adds constraints on subsequent moves. (b) The frequency distribution of branching factors at different depths, corresponding to puzzle development, showing a characteristic three-phase progression. At depth 1, any piece may be selected, resulting in a nearly uniform distribution as each piece contributes a distinct branching factor. As depth increases, the distribution becomes increasingly centered around a lower median value, reflecting the narrowing of available moves. By the final depth, the variance collapses and the branching factor converges to zero, indicating the placement of the last piece and the dominance of unproductive states (the low density of solutions) in the Soma Cube puzzle.}
  \label{fig:full_bf}
\end{figure}

Among the many combinatorial properties of (the state space of) this puzzle, we focus on analyzing its branching factors, which provide a broad overview of the complexity of both the configuration graph $G_P$ and the search space graphs $G_S$. The branching factor measures the number of possible moves available from any given configuration when solving the puzzle. In trivial puzzles, choices at each step are limited, whereas more complex puzzles present multiple valid options, expanding the search space. In the full configuration graph $G_P$ of the Soma Cube puzzle, we find (Fig.~\ref{fig:full_bf}(b)) that the branching factor varies significantly even at depth 1 (where depth measures how many of the seven pieces have been placed, from 0 to 7), with some states having far more successor configurations than others. This variation highlights the structural asymmetry of the state space, which we later exploit to develop efficient solving strategies.

Because of the size of the search space, we employ a sampling approach to compute the branching factor of the Soma Cube puzzle at varying depths using $G_P$ (Fig.~\ref{fig:full_bf}(a); App. \ref{sec:SI3}). The average branching factor is approximately 29.28 (95\% CI: [27.84, 30.72]) \footnote{This calculation compares the number of nodes created in each iteration to the previous iteration. It includes all nodes in the search tree except solution leaves, which would have branching factor 0.} and branching factors decrease monotonically as depth increases. Fig.~\ref{fig:full_bf}(b) visualizes the distribution of branching factors at each depth, showing an amorphous combinatorial distribution at low depth, high variance with an almost-smooth unimodal distribution at low depths, followed by a narrowing of variance (number of available placements) at higher depth. This, overall, reveals the funnel-like topology of the puzzle towards solutions where the width of the state-space graph contracts significantly as depth increases. This combinatorial structure is in contrast with some other puzzles like the Rubik's Cube or Sliding 8-Puzzle, which maintain an almost-constant branching factor throughout, or chess, which does not seem to exhibit monotonic behavior (App. \ref{sec:SI2}). (However, random chess play does show a bell-shaped distribution of branching factors in its midgame, paralleling some of the dynamics we observe in the Soma Cube at middle depths (App. \ref{sec:SI4}).) We will revisit how these mid-game constraints support the emergence of landmarks in Sec.~\ref{sec:landmarks}.

In the following, we consider various algorithmic approaches to solving the Soma Cube. We start with depth-first search (DFS) with backtracking and progressively incorporate value ordering, variable ordering, pruning, and landmark-based heuristics. These algorithms abstract various natural strategies humans employ, guided by the presence of the physical artifact.

Like many puzzles, the Soma Cube is both a logical problem and a physical device for solving it. The tangible pieces restrict how solvers traverse the search space and, at the same time, reveal structure that supports efficient strategies. These observations lead us to propose the following \emph{Principle of Materiality}:
\begin{quotation}
\noindent \textbf{Principle of Materiality}: the physical embodiment of a logical problem can alter the geometry of the logical search space in physically-derived ways that lead to a more efficient search for its solution.
\end{quotation}
In particular, while it is understood \cite{Chase1973, Zhang1997-qi, dreyerEtAlFeinerman} that physical puzzles exploit perception --- visual and tactile --- to distribute problem-solving, the Principle of Materiality posits an additional benefit, in that  physically embodied puzzles can effectively modify the \emph{logical} structure of the search space.
\emph{The Principle of Materiality seeks to explain when, why, and how a formally identical yet computationally difficult logical puzzle can become far more tractable when instantiated as a physical artifact through a formal change in the logical space.}

\section{Comparing the efficiency of algorithms simulating physical constraints}
In this section, we present our quantitative results on the impact of different heuristics on search efficiency, summarized in Fig. \ref{fig:big_comparison}. In Section~\ref{sec:cognitive}, we revisit these algorithms from a cognitive perspective, examining how their efficiency relates to human problem-solving strategies.

To quantify the efficiency of different search algorithms, we focus on the effective branching factor \( b^{*} \) (Eq. \eqref{eqn:eff_bf}): the average number of children per node generated during search, 
\begin{equation} \label{eqn:eff_bf}
\sum_{d=0}^{6} (b^{*})^d = N
\end{equation}
where $d$ stands for depth and $N$ represents the total number of nodes produced during search. This metric is widely used to evaluate the effectiveness of heuristics in improving search performance \cite{russell2010artificial}.

\subsection{Depth-first search with backtracking}
Our first and simplest algorithm is 
a randomized DFS with backtracking. This corresponds to a scenario where the solver does not use sophisticated strategies that exploit cube symmetries or piece constraints (which we introduce as optimizations below), but can still avoid repeating unproductive paths without using elaborate learning techniques. 

\begin{figure}[h!]
        \centering
        \includegraphics[width=\linewidth]{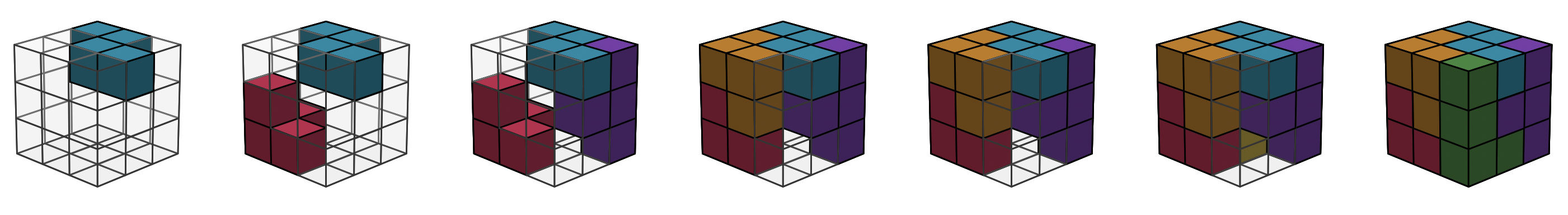}  
    \caption{{ \label{fig:soln_randomized} A solution path found by randomized DFS. Pieces are placed in any empty cell regardless of contiguity.}}
\end{figure}

At each iteration, the solver randomly selects from available empty positions and explores different piece orientations \footnote{Although randomization can be applied to piece selection order, orientation sequences, position selection order, or any combination thereof, for simplicity our implementation randomizes position selection while maintaining a fixed piece and orientation order.}. Upon exhausting valid moves, the algorithm backtracks to the previous valid configuration and explores alternative paths. The solver implements no adjacency requirements between pieces, allowing placement at any unoccupied position within the cube, as shown in Fig.~\ref{fig:soln_randomized}.
\subsection{Contiguous variable ordering}
\begin{figure}[h!]
  \centering

  \subfloat[A solution path found through variable ordering by binding, allowing only contiguous moves.\label{fig:value_moves}]{
    \includegraphics[width=\linewidth]{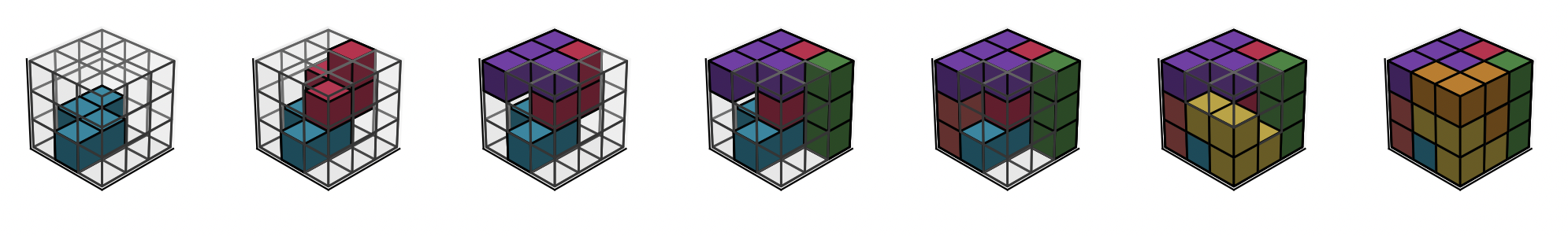}
  }

  \vspace{1em}

  \subfloat[A solution path found through layer-based variable ordering.\label{fig:layer_moves}]{
    \includegraphics[width=\linewidth]{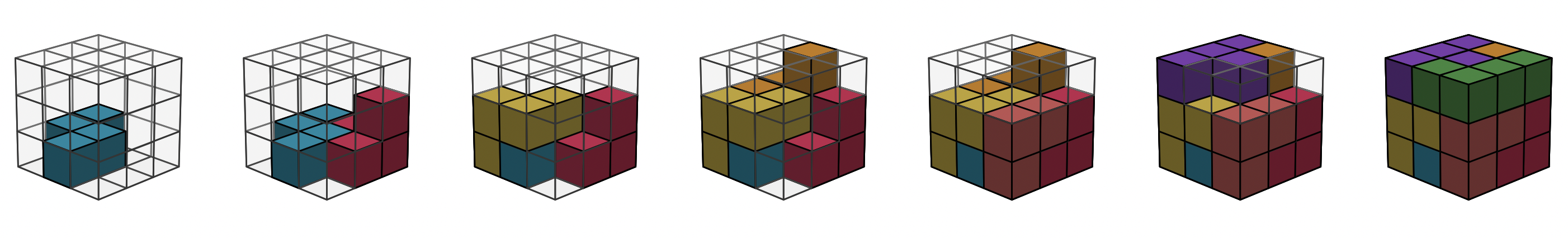}
  }

  \caption{\label{fig:two_paths} Comparison of two different solution paths.}
\end{figure}

For the second solver, we implement cell-ordering, where we require pieces to be placed in positions that cover coordinates in a fixed sequence deterministically.
The solver starts from the (0,0,0) position at the bottom corner of the cube and locates pieces contiguously, as shown in Fig.~\ref{fig:two_paths}(a). 

We then apply layer-based variable ordering, a secondary ordering that prioritizes maximizing coverage of the bottom layer before progressing to higher levels of the Soma Cube during the search, as shown in Fig.~\ref{fig:two_paths}(b). This approach respects the ``gravity constraints'' of the physical puzzle, simulating how it could be solved on a flat surface in the physical world.

\begin{figure}[h!]
  \centering

  \subfloat[Distribution of branching factors at different depths (1--6) for the randomized DFS solver, and the associated backtrack distribution.\label{fig:bf_rand}]{
    \includegraphics[width=\columnwidth]{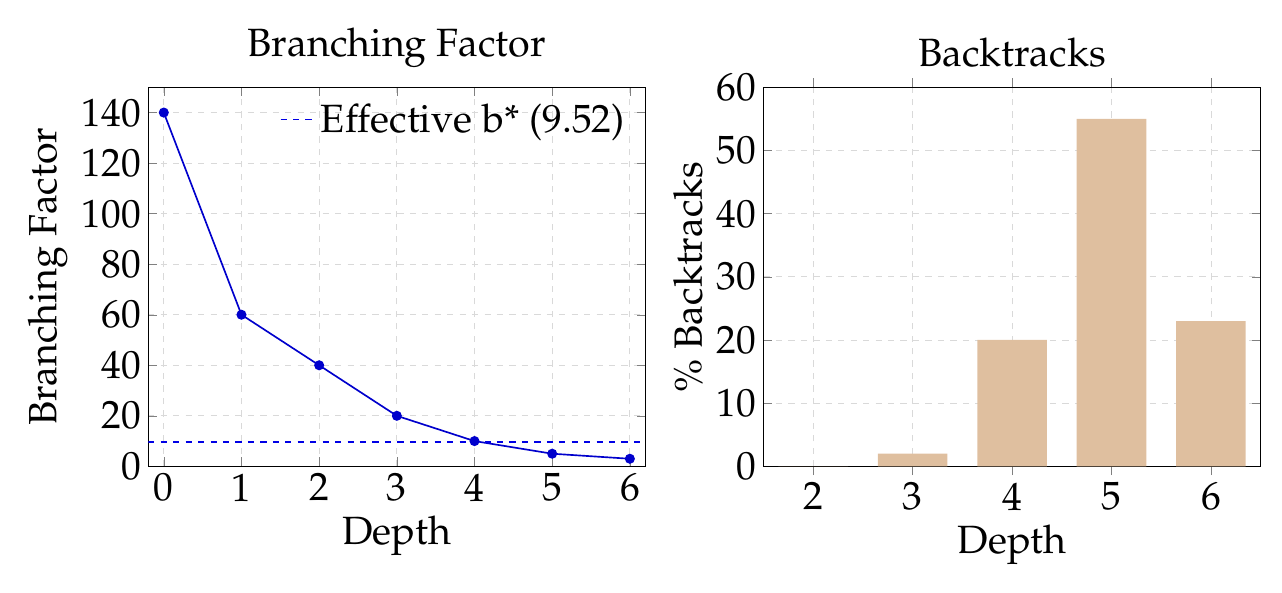}
  }

  \vspace{0.8em}

  \subfloat[Distribution of branching factors at different depths (1--6) for the cell-ordered DFS solver, and the associated backtrack distribution.\label{fig:bf_cell}]{
    \includegraphics[width=\columnwidth]{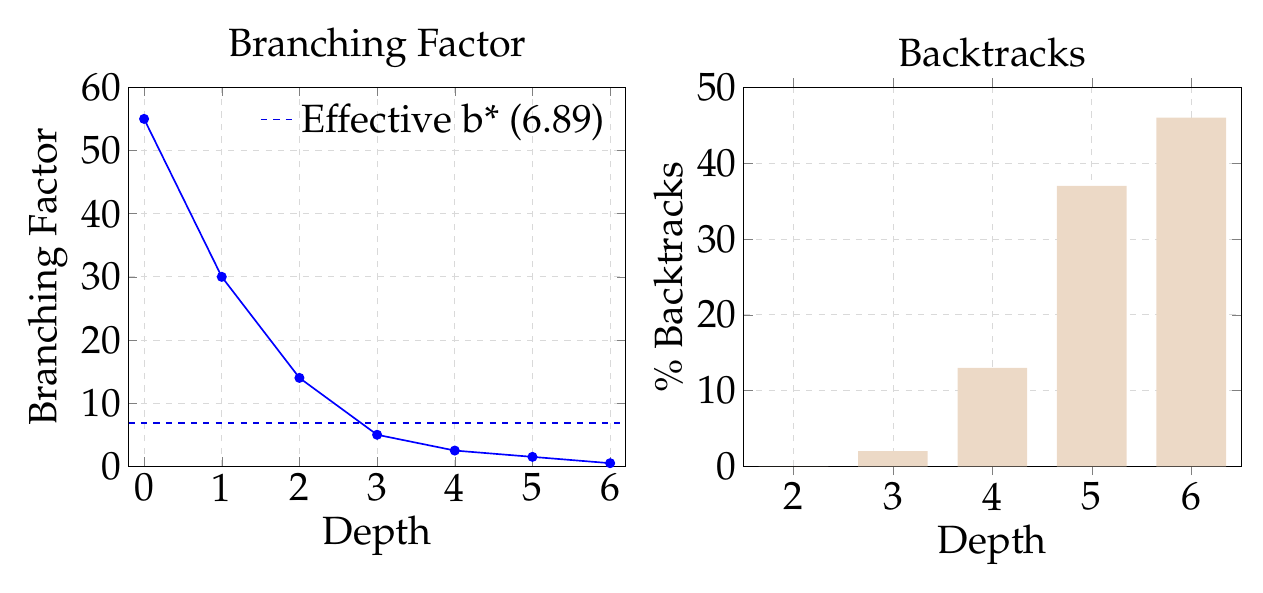}
  }

  \caption{Branching factors and backtrack distributions in the Soma Cube puzzle. For both the randomized DFS and the cell-ordered DFS the branching factor declines exponentially with depth. Randomized DFS tends to backtrack sooner than cell-ordered DFS because an incorrect random piece placement forces a retreat to an earlier configuration. In contrast, the cell-ordered algorithm makes more structured choices, leading to minimal corrections earlier and backtracking primarily during the final piece placement.}
  \label{fig:backtrack}
\end{figure}

The efficiency of these solvers can be understood through their branching factor trends across depths and their impact on the distribution of backtracking events in the search tree (Fig.~\ref{fig:backtrack}). Randomized DFS without contiguous variable ordering (Fig.~\ref{fig:backtrack}(a)) exhibits a heavily front-loaded backtracking pattern, as arbitrary placements often lead to early failures, limiting deeper exploration in a search space with many cycles. However, DFS with cell-ordered variable ordering (Fig.~\ref{fig:backtrack}(b)) enforces adjacency constraints. The difference in backtracking distributions suggests early mistakes have a disproportionately negative impact on search efficiency, whereas deferring errors in a structured manner inspired by the physics of the puzzle enables a more thorough exploration of promising regions in the state space.

\subsection{Dynamic pruning}
Pruning improves search efficiency by eliminating regions of the state space that cannot lead to a solution \cite{russell2010artificial}. Here we implement dynamic pruning, where pruning decisions are made adaptively during the search, based on the current puzzle configuration. 
Our algorithm identifies 
small ``voids'' ---disconnected empty regions of size 1 or 2--- which cannot be occupied by any remaining piece (since all pieces have at least 3 cubes), and prunes and configuration with such voids.

\subsection{Most constrained variable ordering (MCV)} In MCV ordering, we choose the next piece to place by identifying the empty cell with the fewest valid placement options, as determined by the current state of the puzzle. This strategy examines each unassigned cell's possible moves, each defined by a legal combination of piece and orientation, and selects the cell where the number of legal pairs is minimal.

\subsection{Landmarks and hierarchies} \label{sec:landmarks} 
\begin{figure}[h!]

        \centering
        \includegraphics[width=\linewidth]{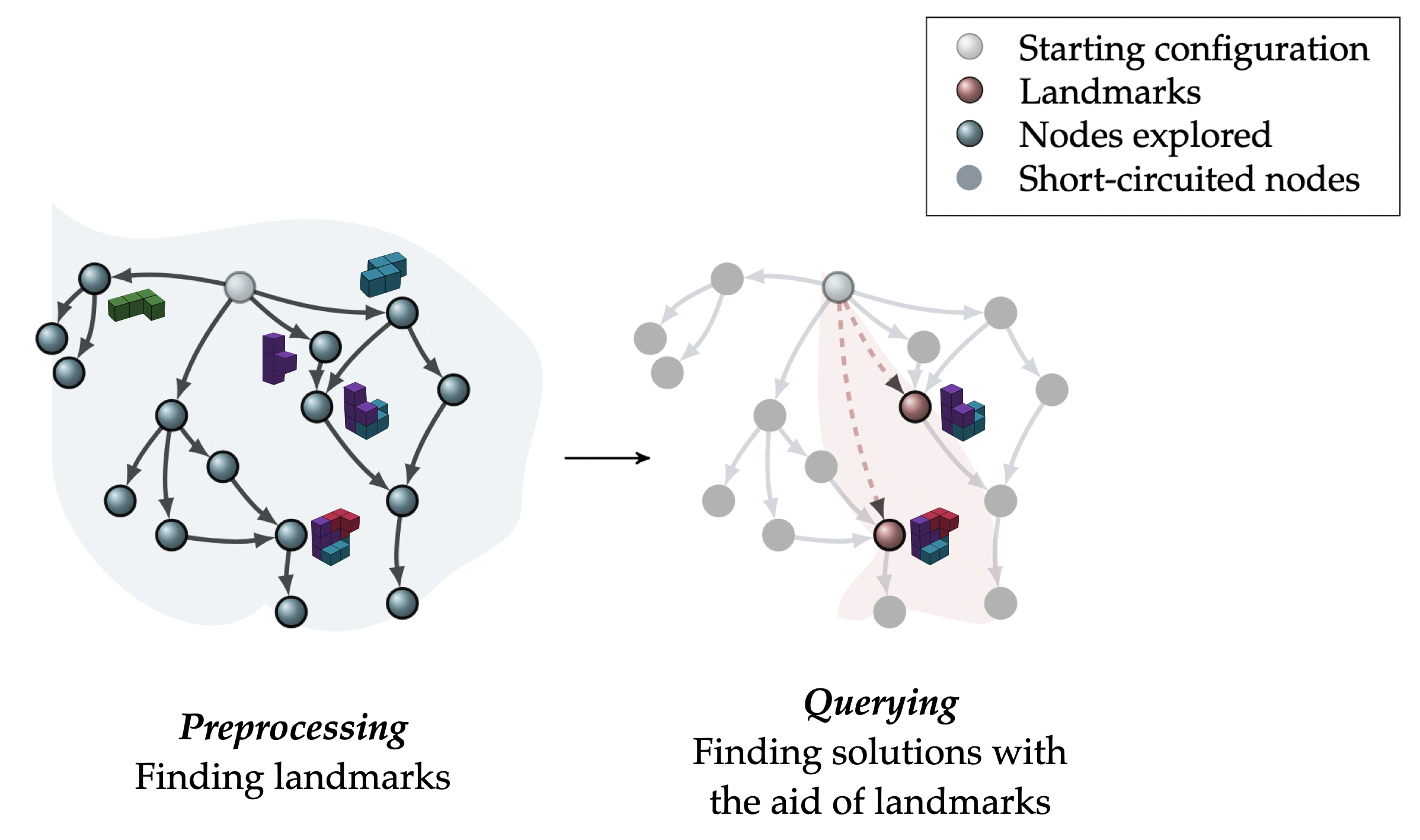}  
        \caption{ \label{fig:cartoon_preprocessing} Comparison of search space exploration strategies for the Soma Cube puzzle.  Left: Standard search space represented by nodes (configurations) and edges (legal moves), with the blue shaded region illustrating the extent of explored states by DFS. Right: Landmark-augmented search incorporating frequently occurring partial configurations as high-importance nodes, with dashed edges representing precomputed successful paths to landmarks and the pink shaded region showing the reduced search space.}   
\end{figure}

Modern graph navigation algorithms like ALT and Contraction Hierarchies accelerate route-finding on graphs by using landmarks and organizing nodes in importance-based hierarchies \cite{goldberg2005computing,Geisberger}. In our implementation for the Soma Cube, we evaluate partial configurations by determining the number of complete solutions that can be derived from each. Configurations that yield a substantial number of full solutions are classified as high-importance landmarks, whereas those that do not lie on any solution paths are designated anti-landmarks (or dead states). We augment the solver so that it can jump directly to designated high-importance landmark states and then resume the search (e.g. via DFS) from those landmarks (Fig.~\ref{fig:cartoon_preprocessing}).

\begin{figure}[h!]
        \centering
    \includegraphics[width=\columnwidth]{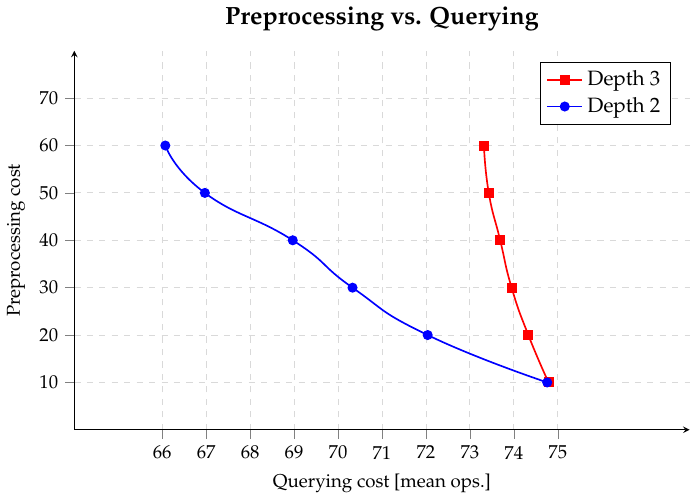}  \caption{ \label{fig:tradeoff} {Trade-off between preprocessing time (time spent identifying and storing landmark configurations) and querying time to reach a solution (actual puzzle-solving time) for varying numbers of landmarks obtained at depth 2 and depth 3 via the cell-ordered DFS solver. At all depths, a reduction in querying cost (increased dependence on look-up tables) is achieved by an earlier increase in preprocessing time (via comprehensive search for landmark configurations). This trade-off is greatest at the puzzle depth where landmarks are most important: a sweet spot defined by a high branching factor and non-trivial constraints on piece placement (Depth 2). As branching factors decline querying provides a diminishing marginal benefit (Depth 3).}}    
\end{figure}

The landmark-based search algorithm is divided into two distinct phases: preprocessing and querying. During preprocessing, a variable number of landmark configurations and all encountered anti-landmarks are identified and stored. In the subsequent querying phase, these precomputed landmarks are invoked to augment the search during the actual puzzle-solving process. There is a trade-off between these two phases (Fig.~\ref{fig:tradeoff}): as the number of landmarks increases, the preprocessing time correspondingly rises, while the querying time decreases.

\begin{figure}[t!]

        \centering
        \includegraphics[width=\linewidth]{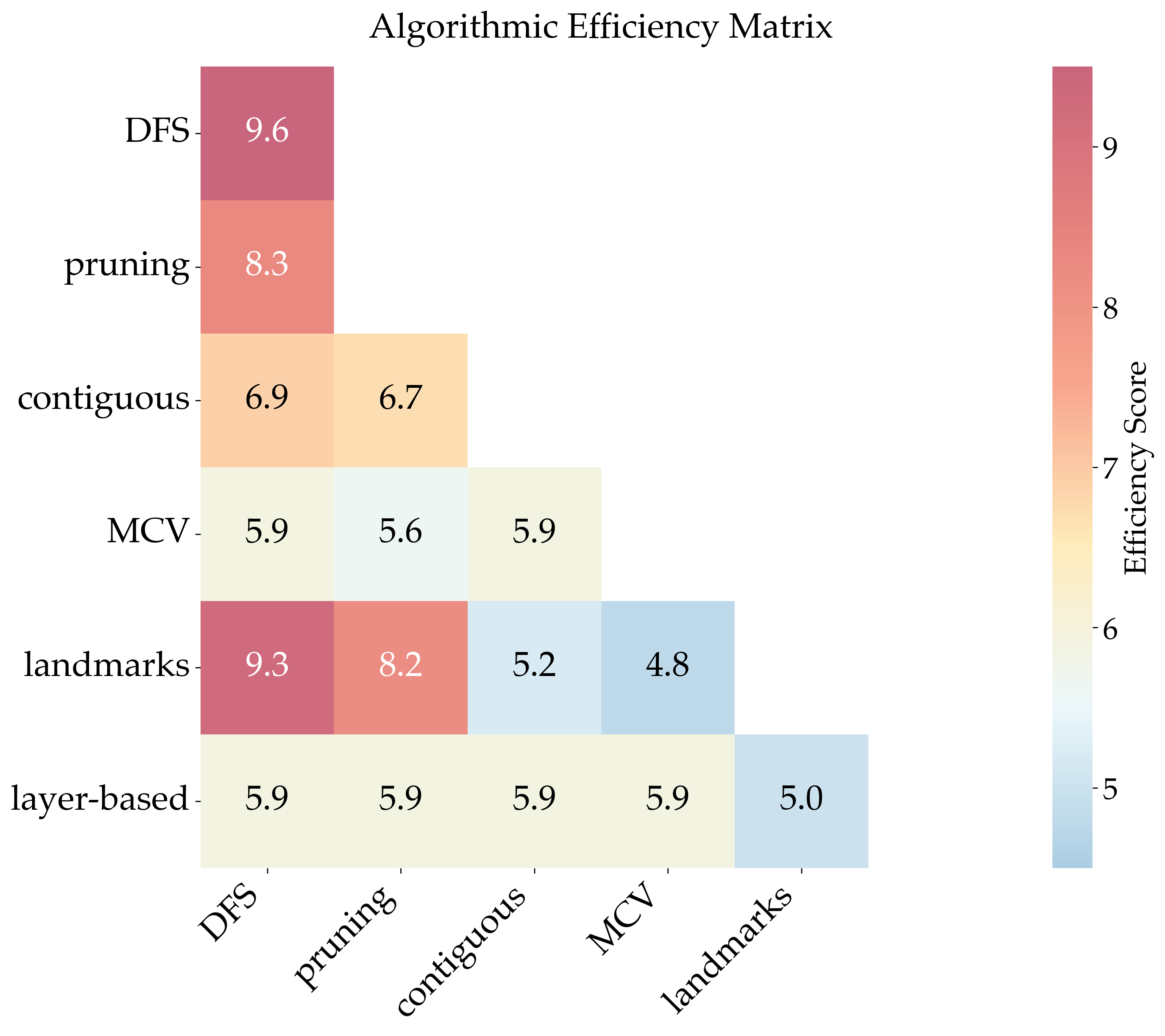}  
        \caption{\label{fig:big_comparison} {\small{Effective branching factors when combining different search strategies for the Soma Cube puzzle.  Values are rounded to a single decimal place.  The heat map shows the effects of structured ordering, pruning, and guidance via landmarks. The largest improvements (above the baseline of randomized DFS) are associated with adding pruning and landmarks. Both of these contract the logical search tree, one by ruling out certain dead ends, and the other by additionally caching shortest paths into lookup tables.}}} 
\end{figure}

\subsection{Quantitative comparison} Given these algorithms, our main result is summarized in Fig.~\ref{fig:big_comparison}, which compares the effective branching factors across different search strategies. Among these approaches, the landmark-based solver achieves the greatest complexity reduction, as preprocessing to identify high- versus low-importance states enables the solver to bypass large portions of the search space. The identification of anti-landmarks avoids dead ends, allowing the solver to sidestep extensive backtracking \footnote{For this figure we derive landmark values by computing upper bounds on the branching factor using anti-landmark configurations. Because landmark sets vary between solvers, these values serve as conservative estimates of the effective branching factors for solvers augmented with the landmark approach.}.

\section{Connecting physically-constrained algorithms to cognitive strategies}
\label{sec:cognitive}
Each algorithm we designed to enhance search within the logical space of the Soma Cube attempts to model a physical aspect of the puzzle. In this section, we ground each algorithm in empirical literature on cognition with physical artifacts, connecting physical and cognitive principles to logical search and constraints. 

\subsection{DFS with backtracking}
\emph{DFS with backtracking implements a physical exclusion principle}.
Solving the $k$-SAT formula is challenging in part because the puzzle’s 27 atomic cubes combine to form 7 compound shapes arranged in 3D space. Rotation of a shape in physical space solves without effort the harder task of rotating 3--4 bound cubes in the mind and tracking their respective coordinates. Mental rotation is a well-attested visuospatial challenge \cite{Shepard1971-zc}. Indeed, the difficulty of rotation was one of the primary arguments in favor of the extended mind \cite{Clark1998-vi}.  The simpler problem of 2D rotation forms the basis of the Tetris puzzle, where the ability to quickly rotate and place pieces is a characteristic of expert play \cite{Lindstedt2019-rw}. Blindfold Tetris, a more challenging variant that relies heavily on auditory cues, further exemplifies this difficulty. Our DFS with backtracking implementation simulates the steric constraints of objects to eliminate non-viable configurations: piece positions that would otherwise require thousands of binary evaluations to discard in a purely abstract logical search.

\subsection{Contiguous variable ordering}
\emph{Variable ordering implements local space filling}. 
Local growth of Soma Cube solutions (piece by piece assembly) ensures that empty spaces or voids in partial solutions are excluded. The physical pieces are natural void detectors which reveal ordinal relations (compatible physical sequences) directly to perception that would otherwise need to be calculated based on inferred intervals (perceived distances in logical space). The ability to perceive spaces or intervals is harder than ordinal depth perception and this effect is amplified when evaluated within continuous manifolds \cite{Norman1998-ra}. Overall humans are  poor at estimating 3D metric structures and performance varies across rotation in viewing angles \cite{Todd2003-ta}. 

\emph{Gravitational, layer-based variable ordering implements layered assemblies}. The difficulty of manipulating pieces to achieve tesselation can be assisted by using a fixed surface. By implementing a local growth rule constrained to layer-by-layer assembly, the ``affordances'' of gravity are simulated to assist the dexterous placement of pieces. A similar principle has been discovered for ``Robot-Grippers'' where the manipulation capabilities of a robotic hand are augmented by a high friction two dimensional surface \cite{Lu2024-rj}. 
\subsection{Dynamic pruning and most constrained variable ordering}
\emph{Dynamic pruning and most constrained variable ordering (MCV) capture the memory of common preferred patterns and procedures using physical pieces}. Board games including chess, as well as simple regular lattice games like Tic-Tac-Toe, have been studied extensively as models for working memory \cite{Galton1894-wn, Gobet1996-vv, Robbins1996-fl, Zhang1997-qi}. Expertise in these cases consists in sustained practice to form affective representations of sensible configurations, from ``opening moves'' through to ``endgame moves'' \cite{Ericsson1994-zw, Mechner2010-bx}. In addition, master chess players prune inferior moves based on fewer and more expansive board scans than amateurs \cite{Blignaut2008-wi}. 
Dynamic pruning and MCV are surrogates of physical scaffolds for reducing search spaces: MCV imposes the ``opening game'' rule of starting with corner pieces that impose a maximum of constraints, whereas pruning eliminates placements along search paths that fail to find solutions. 

\subsection{Landmarks and hierarchies}
\emph{Landmarks capture expert chunking and related cognitive speed-bumps}. 
The aggregation of pieces into larger coherent aggregates as a function of repeat exposure, or chunking \cite{Chase1973-ch}, is a means of achieving significant ``rank-reduction'' in data structures,  allowing for faster perceptual discrimination \cite{Gobet1996-vv}, leaps in performance \cite{Gray2017-xu}, efficient analogy-making \cite{Hofstadter2001-eq}, and improved speed and accuracy \cite{Wu2023-ep}. Landmark algorithms simulate chunking by replacing a multiplicity of possible paths with ``precomputed'' paths through high frequency assemblies. In this way ``preprocessing time'' is used to significantly reduce ``querying time'', what has been called a ``cognitive speed bump'' \cite{Lindstedt2020-sk}, which, in theoretical computer science, can be described as a ``contraction hierarchy'' in the search space graph \cite{Geisberger}.

\section{Discussion}

A physical puzzle both poses a logical problem and offers a mechanical tool for the solution of that problem. We suggest that the quality of a puzzle can be measured by how well the tool solves the problem, i.\,e., how effectively the physical design guides you toward the solution. A well-known example is the Rubik's Cube, which is a tangible representation of a Cayley graph where the cube’s face rotations act as the formal generators of the underlying group, thereby mapping its entire configuration space onto a physical artifact \cite{cooperman1991applications}. Solving the cube optimally requires 
finding short paths in an exponentially large graph, 
a problem that is computationally difficult, yet routinely solved through the applications of manipulation algorithms \cite{demaine2011algorithms}. In this study, we reconceptualize the challenge of the Soma Cube \cite{goodman2019soma} as a search problem and explain why when mapped to a physical artifact the search problem can be solved rather effortlessly. We call this transformation of a challenging logical problem into an easily solvable one through the use of a physical artifact, the Principle of Materiality. This paper examines how the Principle of Materiality operates by mapping physical phenomena to algorithmic processes. Each algorithm emulates a specific physical constraint (rigidity, exclusion, rotation) that effectively reduces the size of the search tree. Hence physical puzzles transform extensive logical problems into contracted search spaces that are cognitively manageable for humans, under constraints of memory and inference. Each of the algorithmic restrictions that we observe can be mapped onto a concept in cognitive science, and this study recapitulates in a purely computational setting the importance of chunking (landmarks and hierarchies), spatial constraints and the role of physical laws including gravity in assisting search (variable ordering), and the anticipatory elimination of nonviable paths (pruning). 

The use of games and puzzles have a long history in mathematics \cite{Rouse-Ball1939-jv}, computer science \cite{Conway1976-ro},  cognitive science \cite{Ericsson1993-cs}, economics \cite{von-Neumann1953-ho} philosophy \cite{Suits1990-de} and history \cite{Huizinga1970-xs}. In one of the canonical texts on puzzles in mathematics, W.~W.~Rouse Ball organizes puzzles into problems of arithmetic, geometry, mechanics, and search (in mazes). All physical puzzles are relegated to a chapter on ``Some Miscellaneous Questions'' as they all deploy some messy combination of arithmetic, geometry, and mechanics, in novel ways. 
This paper aims to systematize and formalize some of these creative interplays. 

Physical puzzles make connection to ideas of an ``Extended Mind'' \cite{Clark1998-vi}, ``Cognitive Artifacts'' \cite{Norman2013-bb}, ``Cognition in the Wild'' \cite{Hutchins2014-qa},  ``Distributed Cognition'' \cite{Zhang2006-uw}, and the role of material culture in the evolution of mind \cite{Dennett2018-tn}. They connect  to the idea of a mental ``physics engine'' that augments purely logical  problem-solving \cite{Schwettmann2018-ud} and might be described as outsourced "physical-physics engines". Owing to the simplicity and intuitive apprehension of games and puzzles, they provide a very convenient model system for exploring general cognitive mechanisms \cite{Allen2024-wu}. 

It is important to distinguish that games are typically designed for multiplayer contexts where competition and coordination are paramount \cite{von-Neumann1953-ho}, while puzzles are structured as solitary challenges centered on constraint satisfaction \cite{Rouse-Ball1939-jv}. Future work might further investigate the cognitive and computational complexity mechanisms underlying these differences, potentially inspiring novel designs that bridge the gap between recreational puzzles and strategic gameplay.

These distinctions in design and engagement not only shape how we interact with artifacts in games and puzzles, but also influence the computational strategies used to analyze them. 
Traditional computational approaches to games and puzzles have followed two distinct paths: complexity theory, which analyzes asymptotic results for generalized versions of games (such as $n \times n$ Tic-Tac-Toe) that bear little relevance to human problem-solving, and search algorithm development, which focuses on optimizing tree traversal for specific games like Chess and Go on modern computers. Similarly, cognitive science research has traditionally focused on studying human subjects tackling computational complexity problems that are proven to be hard in the asymptotic regime \cite{Macgregor1996, Reichman2023}.

A critical gap persists at the intersection of complexity theory, algorithm design, cognitive science, and cultural evolution: we lack a unified framework that explains how the structural properties of finite puzzle instances (not just their asymptotic behavior) determine whether they are engaging or intractable for human solvers. This gap has limited our ability to connect formal computational properties with empirically observed patterns in human problem-solving behavior. The present work uses the physical puzzle and its logical implementation to provide the initial groundwork for a theoretical framework bridging this divide.

\begin{acknowledgments}
G.\ K.\ acknowledges Cris Moore and the SFI library for the puzzles. D. C. K and G.\ K.\  are supported by the Templeton World Charity Foundation, Inc. (funder DOI 501100011730) under the grant DOI:\href{https://doi.org/10.54224/20650}{10.54224/20650} no.\,20650 on “Building Diverse Intelligences through Compositionality and Mechanism Design”, D.C.K. is additionally supported by grant no. 81366 from the Robert Wood Johnson Foundation on Using Emergent Engineering for integrating complex systems to achieve an equitable society. G.\ K.\ is additionally supported from J.\ Grochow and R.\ Frongillo startup funds at the University of Colorado Boulder. J.\ A.\ G.\ was partially supported from NSF CAREER award CCF-2047756.
\end{acknowledgments}

\newpage
\textbf{Data availability:} All code and data used in this study are freely available at GitHub (\url{https://github.com/gulcekardes/cognitive_artifacts}).

\appendix

\section{$k$-SAT encoding}
\label{sec:SI1}

The Soma Cube puzzle consists of seven distinct pieces that must be arranged to fill a \(3\times3\times3\) cube without gaps or overlaps. We encode the puzzle as a Boolean satisfiability problem by assigning a Boolean variable to each potential piece placement, so that a satisfying assignment corresponds to a valid assembly.
We begin by noting that each of the seven Soma Cube pieces is defined by a set of unit cube coordinates. For example, one piece may be represented as $
\texttt{z} = \{(0,0,0),\,(1,0,0),\,(1,1,0),\,(2,1,0)\}$.
To account for the different spatial orientations, our code generates all distinct orientations for each piece by applying rotations about the \(x\), \(y\), and \(z\) axes. After rotations, each configuration is normalized by translating it so that the smallest coordinate becomes the anchor. The \(3\times3\times3\) cube is represented by the set of coordinates
$
\{(x,y,z) \mid x,y,z \in \{0,1,2\}\}
$.
For every piece, every orientation, and every possible anchor position, the code computes an adjusted placement (i.\,e., the translated coordinates that the piece occupies). If all cells of a placement fall within the cube, the placement is deemed valid and is assigned a unique Boolean variable. We denote these variables as
$
P_{xyzi},
$
where \((x,y,z)\) (converted to 1-indexed form) indicates the anchor coordinate, and \(i \in \{1,\ldots,7\}\) identifies the piece.
The overall SAT instance is described by a Conjunctive Normal Form formula, \(\Phi\), which consists of the following constraints:
\paragraph{Piece Constraints.}
For each piece \(i\), we must choose exactly one of its valid placements. Let \(\mathcal{P}_i\) be the set of Boolean variables corresponding to the placements of piece \(i\). This constraint is encoded using:
\begin{enumerate}
    \item An \emph{at-least-one} clause:
    \[
    \bigvee_{P \in \mathcal{P}_i} P,
    \]
    which ensures that the piece is placed somewhere.
    \item \emph{At-most-one} clauses, consisting of pairwise constraints:
    \[
    \neg P \vee \neg P', \quad \text{for all distinct } P,P' \in \mathcal{P}_i,
    \]
    which guarantee that the piece is not placed in two different locations simultaneously.
\end{enumerate}
\paragraph{Cell Constraints.}
Similarly, each cell in the cube must be covered by exactly one piece. For each cell \(c\), let \(\mathcal{C}_c\) be the set of Boolean variables corresponding to placements that cover \(c\). The constraint is again encoded as:
\begin{enumerate}
    \item An \emph{at-least-one} clause:
    \[
    \bigvee_{P \in \mathcal{C}_c} P,
    \]
    ensuring that the cell is covered.
    \item \emph{At-most-one} clauses:
    \[
    \neg P \vee \neg P', \quad \text{for all distinct } P,P' \in \mathcal{C}_c,
    \]
    preventing more than one piece from occupying the same cell.
\end{enumerate}
Thus, the overall formula \(\Phi\) is a conjunction of all these clauses. 
The generated CNF formula \(\Phi\) is then passed to a SAT solver (specifically, \texttt{Minisat22} from the PySAT library). The solver searches for an assignment of truth values to the \(P_{xyzi}\) variables such that \(\Phi\) evaluates to true. A solution to the SAT instance directly corresponds to a valid placement of all pieces in the cube.

\paragraph{Solution Enumeration.}
In addition to finding one solution, our code employs a blocking clause method to enumerate all solutions. After a solution is found, a blocking clause (the negation of the current model) is added, and the solver continues to search for other solutions. 

\section{Branching factor}
\label{sec:SI2}
\begin{figure}[H]
        \centering
        \includegraphics[width=\linewidth]{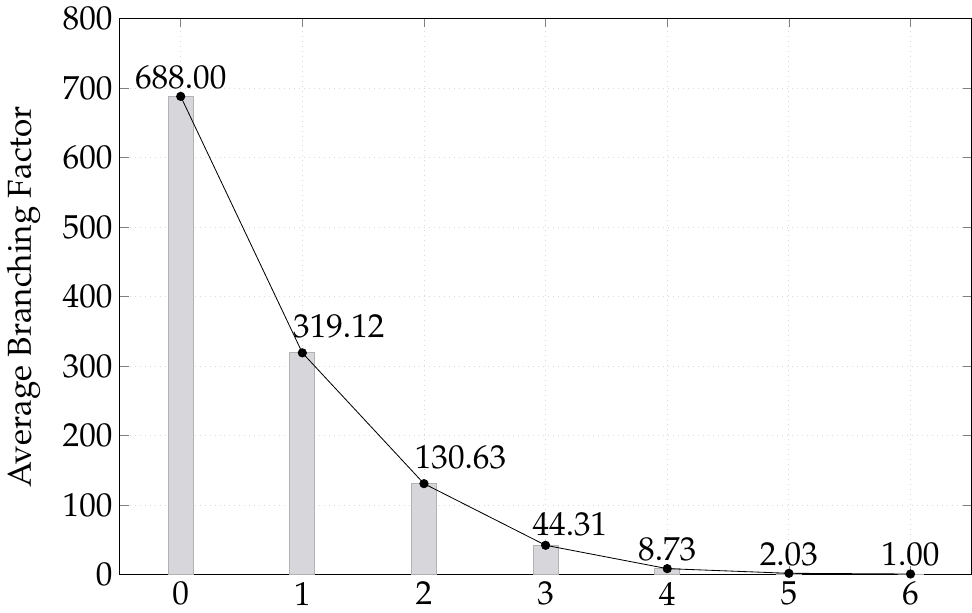}  
        \caption{{\small{Soma Cube branching factors across depths.}}}    

      \label{fig:supp_branching_factors}
\end{figure}

The branching factor of a node in a search tree refers to the number of child nodes it has. In search trees where the number of child nodes varies across different nodes, calculating the average branching factor provides a useful measure of the tree's breadth.
The average branching factor, denoted as $\bar{b}$, can be defined as:
\begin{equation*}
\bar{b} = \frac{1}{D}\sum_{d=1}^{D}\frac{N_d}{N_d + L_d}
\end{equation*}
where $D$ represents the maximum depth of the search tree, $N_d$ stands for the number of non-leaf nodes at depth $d$, and $L_d$ is the count of leaf nodes at depth $d$ that are not solution nodes. The branching factor calculations in the main text of this article include $L_d$ in the formula. However, it is common practice to compute the branching factor for most puzzles with regular search spaces using only nonterminal nodes. For the sake of comparison and reference, we have also provided the results of calculations that follow this convention, see Fig. \ref{fig:supp_branching_factors}.

\section{Branching factors for various puzzles}
\label{sec:SI3}
The analysis of branching factors serves as a fundamental approach for evaluating the complexity inherent in both $G_P$ and $G_S$. In state-space analysis, the diameter (the maximum over all pairs of configurations, of their shortest-path distance) quantifies the worst-case move count between configurations. In a search space, a comparable measure of complexity is the tree’s maximum depth.

\begin{figure}[h!]
        \centering
    \includegraphics[width=\columnwidth]{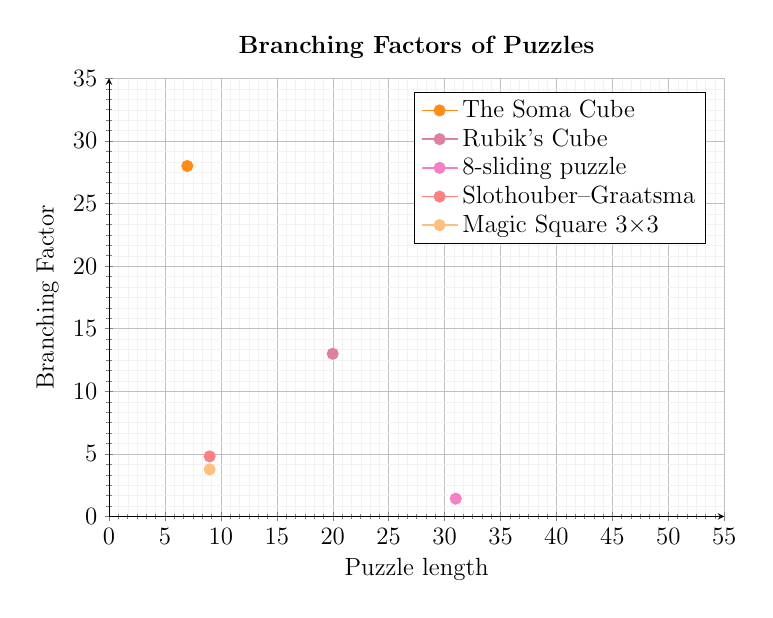}
    \caption{{\small{The relationship between branching factor and depth for five classic puzzles.}}}    
    \label{fig:branching_puzzles}
\end{figure}

Building on this, the worst-case complexity of a tree search is controlled by two parameters, the branching factor and the maximum depth, which determine its time and space costs. A search algorithm must potentially examine the $b^m$ nodes in total, since each of the $m$ levels can have up to $b$ branches, leading to exponential time complexity $O(b^m)$. 
This relationship between $b$ and $m$ demonstrates why problems with high branching factors or large search depths quickly become computationally intractable, even with sophisticated search strategies. 

Fig. \ref{fig:branching_puzzles} displays a taxonomic map of puzzles, organized by their branching factor patterns and depth characteristics, where depth is defined as the minimum number of moves required to solve the puzzle from its worst-case position. For all puzzles except Rubik's Cube, data was generated using our GitHub code \cite{kardes_soma_2025}, while the branching factor information for Rubik's Cube was obtained from \cite{rokicki2010god}.

\begin{figure}[h!]
        \centering
        \includegraphics[width=\linewidth]{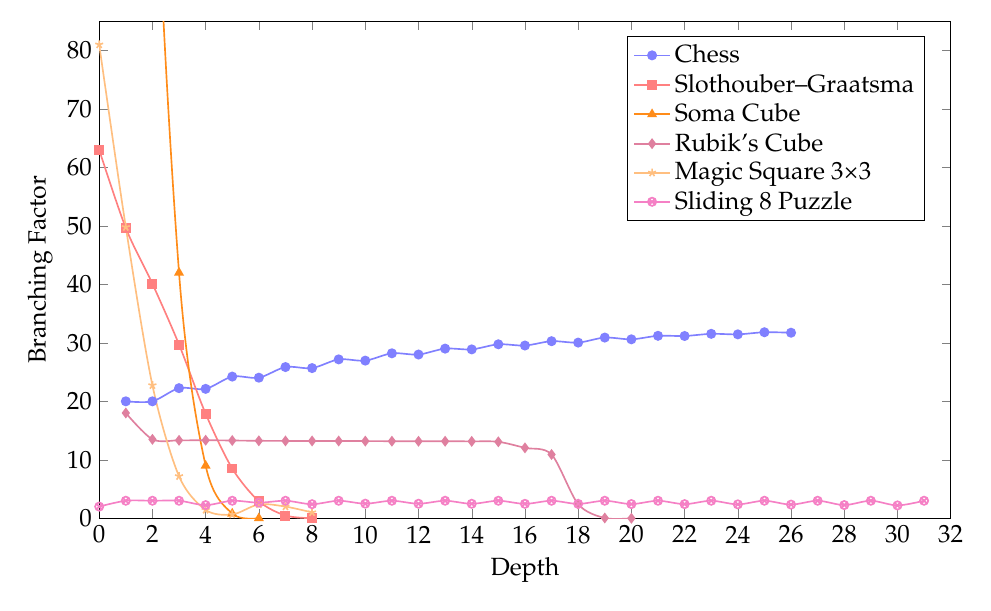}  
        \caption{Branching-factor evolution across multiple puzzles. Each curve shows how the average branching factor changes with depth in the puzzle’s solution process. Chess exhibits a non-monotonic branching factor due to piece dynamics. Rubik’s Cube maintains an almost constant branching factor until the puzzle is nearly solved, reflecting its highly regular structure and uniform move set. The 8 puzzle illustrates a sliding-tile puzzle with moderate non-monotonicity as the empty space’s position affects available moves, while still retaining a regular state space similar to Rubik’s Cube. Three dissection puzzles, the \(3\times3\) Magic Square, Slothouber--Graatsma, and the Soma Cube, each exhibit a steady decline in branching factor as pieces are placed, highlighting their forward-directed, space-constrained nature.}  \label{fig:depths_puzzles}
\end{figure}

Fig.~\ref{fig:depths_puzzles} compares branching-factor evolution across depths for chess, Rubik’s Cube, the 8 puzzle, and three dissection puzzles: the 
Magic Square, Slothouber--Graatsma, and the Soma Cube.

As previously mentioned, variations such as changes in branching factors or other combinatorial properties across different puzzle depths could offer valuable insights into puzzle characteristics and inform potential solution strategies. Among the puzzles we examine for comparison and contrast with the Soma Cube, dissection puzzles (including the Soma Cube itself and two additional examples discussed below) feature irregular state spaces and follow a directed solution process where each piece placement progressively restricts future moves, leading to a consistent decrease in the branching factor. In contrast, Rubik’s Cube maintains a nearly constant branching factor until the end, whereas chess and the sliding puzzle exhibit more complicated trends. 

In the case of the sliding puzzle, the branching factor is nearly constant, reflecting the reversible character of its moves, similar to the Rubik's Cube. However, in the case of the sliding puzzle, the branching factor alternates up and down every other move, because of simple constraints on the moves: if you're not allowed to backtrack, then when in a corner there is only 1 choice, on a side there are 2 choices, and in the middle there are 3 choices. After being in a corner, one must move to a side; after being in a side, one must move to a corner or the center; after being in the center, one must move to the side; and (without backtracking) one can never see the sequence center--side--center. This explains the nearly-constant but every-other-turn character of the branching factor of the sliding 8 puzzle. (This phenomenon would not be present in a variant of the sliding-8 puzzle played on a torus, that is, where the sides wrap around, as in that setting every position is like every other position, with a constant non-backtracking branching factor of 3.)

In the case of chess, we see both a monotonically increasing overall trend (in the first 25 rounds of the game we simulated), but also an alternating up-and-down pattern. The monotonically increasing overall trend in the early moves is also reflected in real play \cite{barnes}, though for fewer moves than we see in our random simulations. But the latter pattern---that the second player seems to systematically have fewer available moves than the first, especially in the first few dozen rounds of the game---has also been observed by David Barnes in his analysis of 2.5M real chess games from ChessBase \cite{barnes}. As far as we are aware, there is as yet no explanation for this phenomenon.

\section{Supplementary Information 4}
\label{sec:SI4}
In addition to the Soma Cube puzzle, similar branching-factor distributions appear in other combinatorial games, most notably chess. Although we do not prove that landmarks necessarily arise whenever a bell-shaped distribution occurs in the midgame, we have strong indications that, when such a distribution appears in a relatively unconstrained midgame (as at depth 2 in the Soma Cube, and in chess, see \cite{Chase1973, delaney2018role} which demonstrate differences in midgame position evaluation where chess masters outperform others due to chunking), introducing landmarks can be beneficial.

\begin{figure}[h!]
  \centering

  \subfloat{%
    \includegraphics[width=\columnwidth]{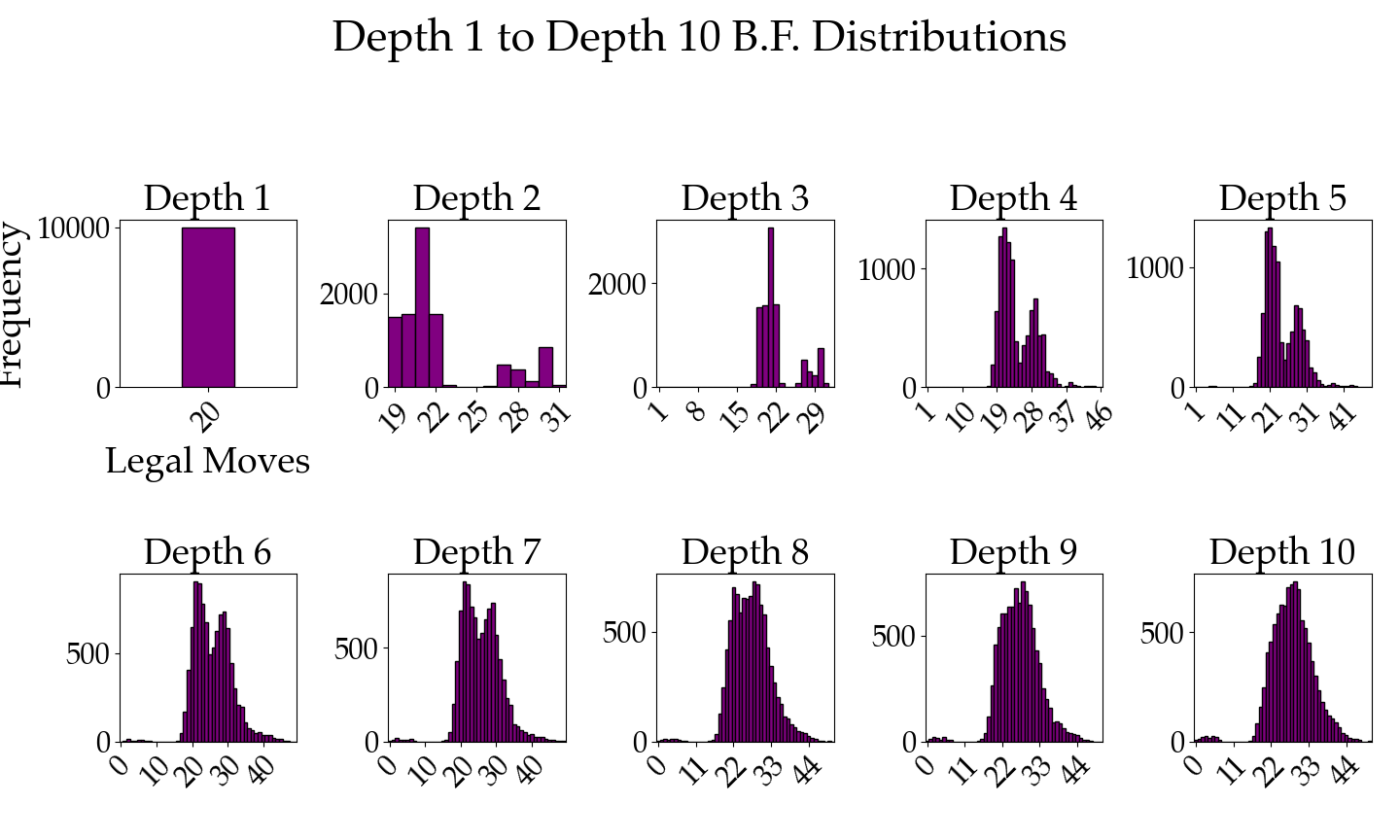}%
  }

  \vspace{0.8em}

  \subfloat{%
    \includegraphics[width=\columnwidth]{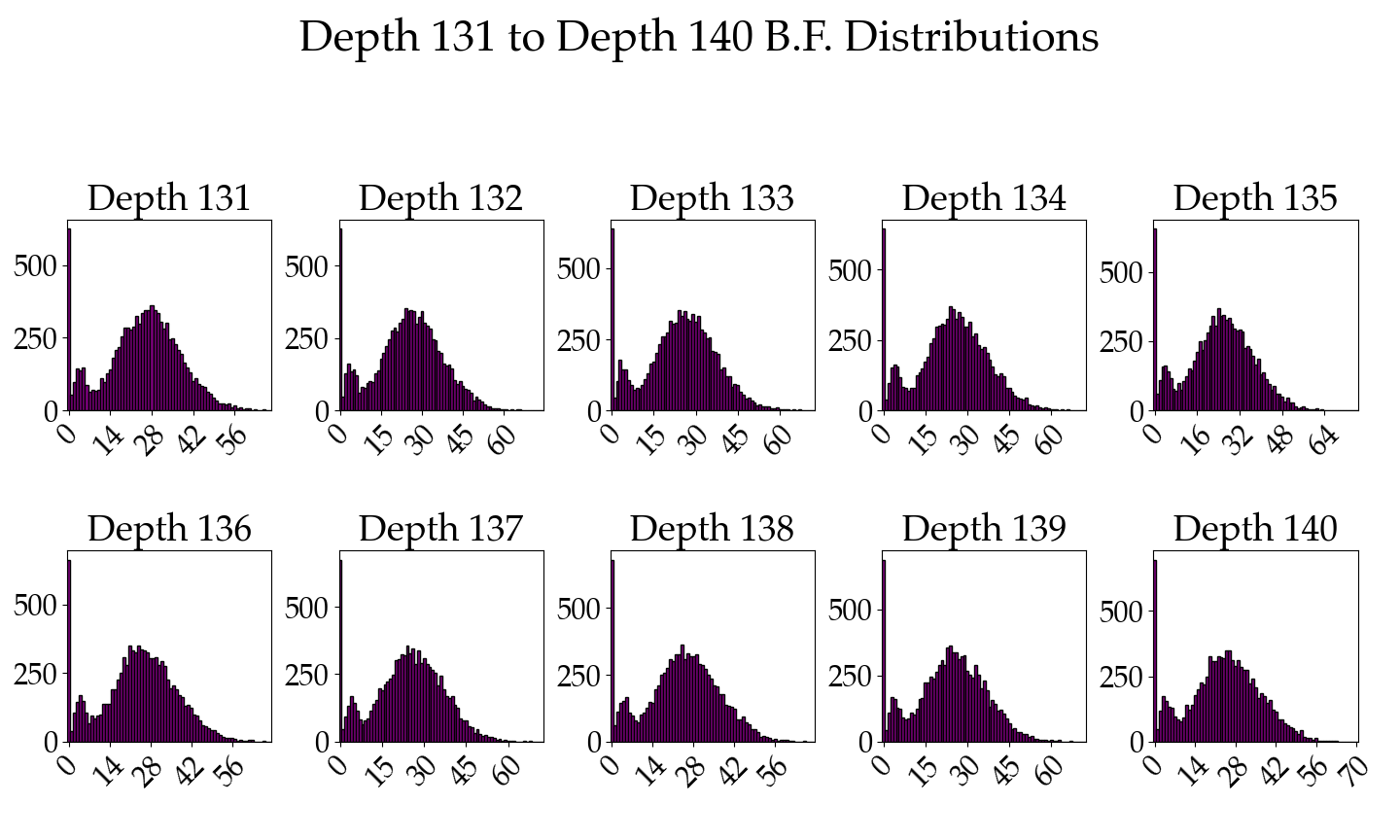}%
  }

  \caption{\label{fig:chessBF}%
    Branching factor distributions in random chess at two distinct ranges of depth.
    \textbf{Top:} Depths 1--10, where the game is in its early stages and the distribution
    tends to be narrower. \textbf{Bottom:} Depths 131--140, where many games have ended
    (spiking at zero moves) or become heavily reduced, causing a shift in the distribution
    and a lower average branching factor.}
\end{figure}

Fig.~\ref{fig:chessBF} demonstrates that, much like in the Soma Cube, the early branching-factor distribution in chess appears relatively uniform, transitioning to a bell-shaped distribution in the midgame. In the later stages, this bell-shaped spread often develops a cluster near zero moves, producing a mixed distribution that ultimately converges to a near-delta distribution in the endgame. 

Here we simulated 10,000 random chess games using the \texttt{python-chess} library \cite{kardes_soma_2025}. Starting from the standard initial position, each ply (half-move) was chosen uniformly at random from all legal moves, and the game was terminated either when maximum depth 140 was reached or when a checkmate/stalemate occurred. We recorded the branching factor at each depth across these simulations and then aggregated the results into histograms, producing Fig.~\ref{fig:chessBF}. This shows how the branching factor evolves from the opening (with relatively uniform move counts), through the midgame (where distributions often become bell-shaped), to the endgame.


\bibliography{apssamp}

\end{document}